\documentclass{article}

    \usepackage[final]{neurips_2023}

\usepackage[utf8]{inputenc} % allow utf-8 input
\usepackage[T1]{fontenc}    % use 8-bit T1 fonts
\usepackage{hyperref}       % hyperlinks
\usepackage{url}            % simple URL typesetting
\usepackage{booktabs}       % professional-quality tables
\usepackage{amsfonts}       % blackboard math symbols
\usepackage{nicefrac}       % compact symbols for 1/2, etc.
\usepackage{microtype}      % microtypography
\usepackage{xcolor}         % colors

\usepackage{bm}
\usepackage{caption}
\usepackage{subcaption}
\usepackage{layouts}
\usepackage{blindtext}
\usepackage{fancyvrb}
\usepackage{amsmath}
\usepackage{amssymb}
\usepackage{cleveref}

\usepackage[textwidth=3cm,textsize=small,disable%
 ]{todonotes}

\newcommand{\Real}{\mathbb{R}}
\newcommand{\R}{\bm{R}}
\newcommand{\Rp}{\bm{R'}}

\newcommand{\Sm}{\bm{S}}
\newcommand{\Smp}{\bm{S'}}

\newcommand{\transp}{\mathsf{T}}
\newcommand{\tr}{\operatorname{tr}}

\title{Towards Measuring Representational Similarity of\\Large Language Models}

\author{%
    Max Klabunde
    \qquad
    Mehdi Ben Amor
    \qquad
    Michael Granitzer
    \qquad
    Florian Lemmerich\\
    University of Passau\\
    \texttt{firstname.lastname@uni-passau.de}
}

\begin{document}

\maketitle

\begin{abstract}
    Understanding the similarity of the numerous released large language models (LLMs) has many uses, e.g., simplifying model selection, detecting illegal model reuse, and advancing our understanding of what makes LLMs perform well.
    In this work, we measure the similarity of representations of a set of LLMs with 7B parameters.
    Our results suggest that some LLMs are substantially different from others.
    We identify challenges of using representational similarity measures that suggest the need of careful study of similarity scores to avoid false conclusions.
\end{abstract}

\section{Introduction}
Numerous large language models (LLMs) with remarkable natural language understanding and reasoning capabilities have been released in recent months \citep{yang_harnessing_2023,zhao_survey_2023}.
However, a comprehensive understanding of the differences between these models beyond architectures and benchmark performances is yet to be established. 
This is partly due to the inherent challenges in LLMs' explainability given their scale, their high demand for computational resources, and the rising trend of proprietary models.

We argue that a thorough understanding of similarities and differences of LLMs is highly desirable:
it may help identify factors that make models perform well, clear up the generalizability of studies of individual LLMs, simplify model selection, enhance our ability to ensure alignment of model behavior with human goals, improve ensembling, benchmark models without labeled data, identify (potentially illegal) model (re)use, and may aid certification of models, which could be required by future regulation of AI.

LLM similarity can be studied from multiple perspectives, including functional similarity, i.e., whether they produce similar outputs, representational similarity, i.e., whether they have similar internal representations, whether they have similar reliance on specific training data, or whether they were trained in a similar manner.
Methods for these perspectives have been proposed in prior work, but often focus on non-sequence models \citep{klabunde_similarity_2023} or do not scale to the size of LLMs \citep{shah_modeldiff_2022}.
In this work, we focus on representational similarity in the last layer as it implies functional similarity, because the final layer has limited options to diverge functionally. 
Additionally, it allows studying similarity of \emph{how} outputs are generated.

Similarity of language models was studied to some extent \citep{wu_similarity_2020,ethayarajh_how_2019}, but these works do not explore similarity of decoder-only models on the scale of recent LLMs, and instead focus on smaller BERT-style models.
As LLMs have developed at break-neck speed, analysis of the similarity of LLMs is generally limited.
Moreover, many novel tools to study the similarity of representations have emerged relatively recently.

In this paper, we aim to make the first steps towards understanding similarity of LLMs in more detail:
\begin{enumerate}
    \item After discussing several options to compare LLMs, we outline how representational similarity measures can be applied to LLMs (Sec.~\ref{sec:comparing_llms}).
    \item We present first empirical results regarding the representational similarity of a set of 7B parameter models, offering a preliminary view into LLM similarity for commonsense reasoning (Winogrande) and code generation (HumanEval) (Sec.~\ref{sec:experiments}).
    \item We identify challenges of gaining a reliable picture of similarity of LLM representations (Sec.~\ref{sec:experiments}).
\end{enumerate}
Our code and data is publicly available (see \Cref{apx:code}).

\noindent\textbf{Related Work.\ \ }
Several works study representations of language models and make implicit comparisons: how contextual they are \citep{ethayarajh_how_2019}, what interpretable concepts can be decoded from them \citep{liu_linguistic_2019}, or how models can communicate via representations \citep{moschella_relative_2023}.
\citet{wu_similarity_2020,abnar_blackbox_2019} explicitly compare representations between models.
However, these works have in common that they do not study the current generation of LLMs, and instead focus on smaller models with different architectures like BERT \citep{devlin_bert_2019} or ELMo \citep{peters_deep_2018}.
Similarity of these models was also studied from a functional perspective \citep{mccoy_berts_2020}.
As an exception, \citet{gurnee_finding_2023} probe the recent Pythia models \citep{biderman_pythia_2023}. 
Concurrent work \citep{yousefi_-context_2023,brown2023understanding} studies representations of modern LLMs.
Performance of LLMs is compared in many benchmarks \citep[e.g.,][]{srivastava_capabilities_of_language_models_2022}.

\section{Comparing Large Language Models}\label{sec:comparing_llms}
In this section, we discuss different ways to study similarity of LLMs and explain representational similarity in more detail, which we focus on in our experiments.

\noindent\textbf{Multitude of Comparison Approaches.\ \ }
There are multiple different approaches to comparing LLMs and measuring their similarity. 
Five approaches are functional similarity, representational similarity, weight similarity, similarity of training data attribution, and procedural similarity.

\emph{Functional similarity} aims to compare the outputs of models \citep{klabunde_similarity_2023}.
A common approach is to compare performance, where performance similarity equates to model similarity.
However, performance alone only gives a partial view of functional similarity: benchmarks represent only a sample of data and models may differ on individual predictions or subgroups of the data, which may impact other functional aspects such as fairness.
Hence, more facets of model function need to be compared for a thorough understanding of functional similarity.

\emph{Representational similarity} compares representations of different layers or models under consideration of their symmetries, i.e., transformations of representations that keep them equivalent such as changing neuron order \citep{klabunde_similarity_2023}.
Studying \emph{weight similarity} is a related perspective requiring consideration of symmetries \citep{wang_understanding_weight_similarity_2022}.
These approaches can identify cases where two models can be functionally similar but produce the same output differently.

\emph{Training data attribution} \citep[e.g.,][]{shah_modeldiff_2022,grosse_studying_2023} is an approach that identifies the most relevant samples of the training data with respect to influencing the model towards a specific output.
From this perspective, models are similar if they have the same relevant training samples for a specific prediction.

Finally, \emph{procedural similarity} is one step removed from the model itself.
It takes the perspective that models are similar if the way they are produced is similar.
This includes similarity of datasets used for training, hyperparameters, and architecture \citep[e.g.,][]{zhao_survey_2023}.

In this work, we focus on one perspective: representational similarity.
In the following, we explain how representational similarity measures can be applied to LLMs.

\noindent\textbf{Representational Similarity.\ \ }
Let $f^{(l)}$ be the model that consists of the first $l$ layers of the model $f$.
Then, given $N$ inputs, their $D$-dimensional representations in layer $l$ are given as $\R:=\R^{(l)}=f^{(l)}(\bm{X}) \in \Real^{N\times D}$, where $f^{(l)}$ is applied row-wise on the inputs $\bm{X}$.
We denote the representations $\R$ is compared to as $\Rp=f^{\prime(l')}(\bm{X})$.
A representational similarity measure $m(\R, \Rp)$ then assigns a similarity (or a dissimilarity) score to two representations.
Although representations may not be identical, they might be seen as equivalent, e.g., if $\R = - \Rp$.
Hence, a measure respecting these symmetries has certain \emph{invariances}.
A measure is invariant to a set of transformations $\Phi$ if $m(\phi(\R), \psi(\Rp)) = m(\R, \Rp)~\forall \phi, \psi \in \Phi$.
We refer to the survey by \citet{klabunde_similarity_2023} for a more detailed introduction.

In this work, we study the representations of the last layer before the final classifier under the invariances to orthogonal transformations (OT), which include rotations and reflections, isotropic scaling (IS), and translation (TR).
We justify this selection with the fact that these representations do not have a privileged basis \citep{elhage_mathematical_2021}, i.e., there is no reason for the basis dimensions (neurons) to have special meaning.
This is because the representations could be arbitrarily rotated by applying the same transformation to the weight matrices that add information to the stream.
The arbitrary rotation validates OT invariance; similarly, IS invariance is useful.
A translation of the representations can affect the outcome of the classification layer for models without a bias, but could also be arbitrarily added by the biases of previous layers.
Hence, we study similarity with and without TR invariance.
The measures we use for OT and IS invariance experiments are Orthogonal Procrustes \citep{ding_grounding_2021}, Aligned Cosine Similarity \citep{hamilton_diachronic_2018}, the norm of the difference of pairwise similarity matrices \citep{yin_dimensionality_2018}, and Jaccard similarity \citep{schumacher_effects_2021,wang_towards_2020}.
For the experiment with OT, IS, and TR invariance, we additionally use RSA \citep{kriegeskorte_representational_2008} and CKA \citep{kornblith_similarity_2019}.
To achieve the desired invariances, we preprocess the representations in some cases by normalizing their scale or centering the columns.
Details are in \Cref{apx:repsim_measures}.

All of these measures make two assumptions that are generally violated with LLMs: (i) the representation of inputs is deterministic, and (ii) all rows of the representations correspond exactly.
With LLMs, the representation of a part of an input sequence depends on representations of earlier parts. These parts are usually sampled if new text is generated, which contradicts (i).
Further, differing tokenization can lead to a different number of tokens for the same input text, and thus to a different number of rows in the representation matrix, which violates (ii).
We can sidestep these problems.
First, we only study the representations of fixed input prompts, which avoids the problem of non-determinism of text generation.
Second, we only compare the representations of the final token in the last layer to avoid the issue of differing tokenization. 
Since these representations are used for the next token prediction, we argue that they have similar meaning across models.
Other solutions to this issue based on aligning differing tokenizations were proposed \citep{liu_linguistic_2019, clark_what_2019}, which enable more fine-grained comparisons at increased computational cost.

\section{Experiment}\label{sec:experiments}

\noindent\textbf{Data.\ \ }
As a first step, we use two datasets from different domains.
Winogrande \citep{sakaguchi_winogrande_2020} is a benchmark aimed at measuring commonsense reasoning abilities by asking a model to fill in a blank in a sentence with binary options.
We use the Winogrande validation set.
Further, we use HumanEval \citep{chen_evaluating_2021}, a code generation benchmark.
Here, prompts consist of a comment that describes the functionality of code that should be generated.
We always feed data in without additional examples (zero-shot).
We show the prompt styles in \Cref{apx:prompts}.

\noindent\textbf{Models.\ \ }
We use a set of 11 freely available LLMs with roughly 7B parameters: RedPajama \citep{togetherai_redpajama_2023}, Bloom \citep{bigscience_bloom_2023}, Falcon \citep{penedo_refinedweb_2023}, Galactica \citep{taylor_galactica_2022}, GPT-J \citep{wang_gpt-j-6b_2021}, Llama \citep{touvron_llama_2023}, MPT \citep{mosaicml_introducing_2023}, OpenLlama \citep{geng_openllama_2023}, OPT \citep{zhang_opt_2022}, Pythia \citep{biderman_pythia_2023}, and StableLM Alpha \citep{stabilityai_stablelm_2023}.
All models are base models without instruction finetunining.
For the code data, we add CodeLlama and CodeLlama-Python \citep{roziere_code_2023}, which are specifically trained on code.
Except for Llama, we use the weights published on Hugging Face.

\subsection{Results}
\begin{figure}
    \centering
    \begin{subfigure}{\textwidth}
        \centering
        \includegraphics[width=0.8\textwidth]{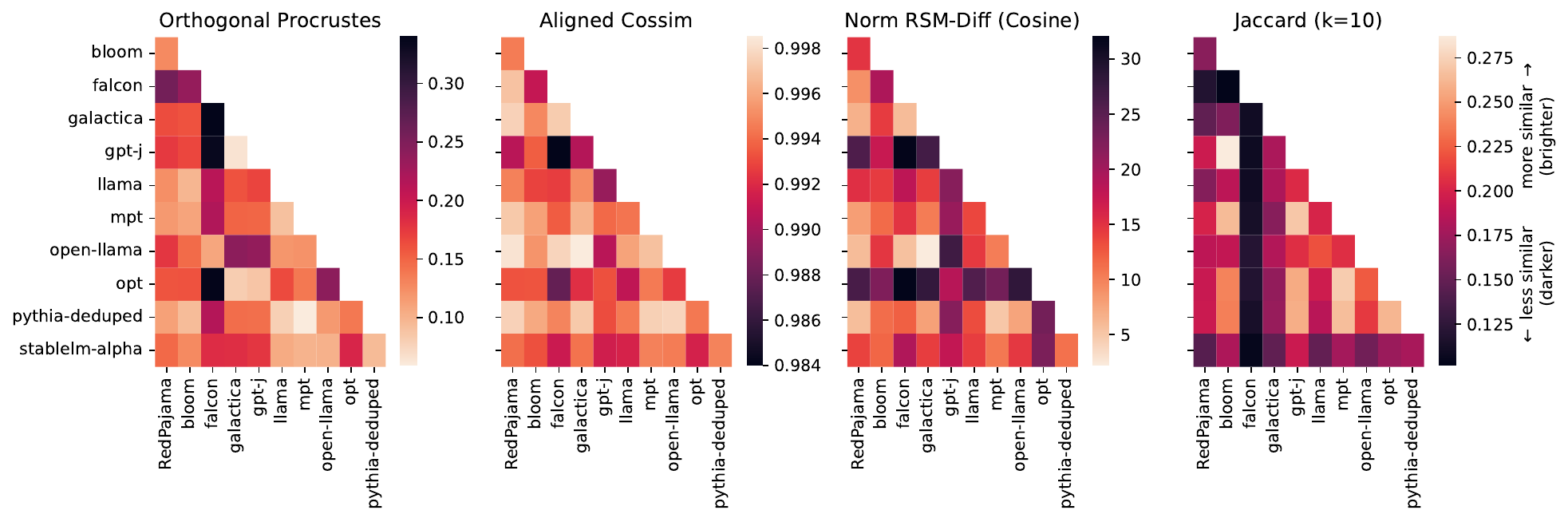}
        \caption{Winogrande.
            Models are not uniformly similar as seen by the non-uniform patterns in the heatmaps. Some models like StableLM Alpha (bottom row) appear relatively dissimilar to all other models.
            Further, the patterns differ between measures. A single measure may not be able to tell the full story (average Spearman $\rho=0.35$).
        }
        \label{fig:repsim_winogrande_otis}
    \end{subfigure}
    \begin{subfigure}{\textwidth}
        \centering
        \includegraphics[width=0.8\textwidth]{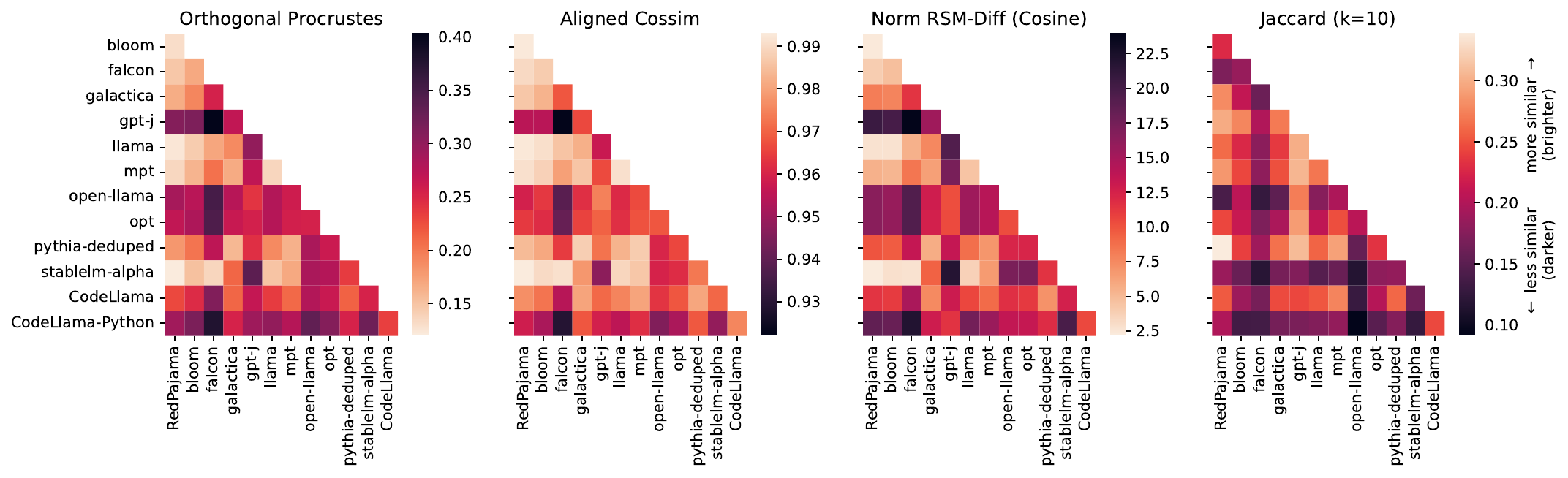}
        \caption{HumanEval.
            As not all models have similar amounts of code in their training data, similarity patterns are expectedly different to Winogrande.
            Of the code-specific models, only the Python variant seems generally dissimilar to the other models.
            Compared to (a), correlation between measures is higher (average $\rho=0.65$).
        }
        \label{fig:repsim_humaneval_otis}
    \end{subfigure}
    \caption{Representational similarity on Winogrande (top) and HumanEval (bottom) with OT and IS invariance.
        Bright colors show the most similar models, dark colors the most dissimilar ones.
    }
    \label{fig:repsim_otis}
\end{figure}

In \Cref{fig:repsim_otis}, we report representational similarity with OT and IS invariance.
Results are similar with OT, TR, and IS invariance, which we show in \Cref{apx:additional_results} due to space constraints.

\noindent\textbf{Significant differences between models.\ \ }
On both datasets, some models have significant differences.
On Winogrande, for example, Falcon stands out for a relatively low similarity by Orthogonal Procrustes and Jaccard similarity.
Similarly, StableLM Alpha seems to be relatively dissimilar to all other models.
On HumanEval, OpenLlama and OPT stand out as dissimilar.

\noindent\textbf{Significant differences between similarity measures.\ \ }
Again looking at Falcon on Winogrande, this model stands out as dissimilar only with two of the four similarity measures.
For both Aligned Cosine Similarity and Norm RSM-Diff, differences to other models are less pronounced.
For these measures, GPT-J and OPT seem more dissimilar instead.
These differences between measures occur despite them sharing the same invariances. It seems that while these measures share the same (high-level) view on representational similarity, finer-grained differences have substantial influence.
Although \citet{ding_grounding_2021} report a similar result for CKA and Orthogonal Procrustes, our results show this is a more wide-spread issue.
Few dimensions with high mean and variance may mask differences in all but these few dimensions for cosine similarity-based measures \citep{timkey_all_2021}.

\noindent\textbf{Similarity is application-dependent.\ \ }
Similarity for one task does not imply similarity for another task:
while GPT-J has a similar Orthogonal Procrustes score to the other models on Winogrande, it stands out as dissimilar on HumanEval.
This difference is expected to some degree as not all models had similar amounts of code in their training data, but still highlights that data dependency is an important factor when making claims about similarity of LLMs.
\\
Additionally, the measures with discrepancies are not the same across datasets: while Orthogonal Procrustes and Norm RSM-Diff differ on Winogrande, they show highly similar patterns for HumanEval.
The average Spearman correlation between heatmaps on Winogrande versus HumanEval is only 0.34.

\noindent\textbf{Difficulty of interpretation.\ \ }
Scores of measures like Orthogonal Procrustes and Norm RSM-Diff that do not have a clear upper bound are difficult to interpret.
In the absence of an interpretable scale, it is unclear whether uniform scores imply all models are equally similar or equally dissimilar.

\section{Conclusions}
We demonstrate measuring representational similarity of LLMs using a set of 7B parameter models.
Representations do not seem to be universal, which may limit generality of study of any single LLM, but boost abilities to detect specific models.
We identify several challenges of using representational similarity measures for measuring LLM similarity: discrepancies between measures, task-dependency, and interpretation.
These challenges provide interesting avenues for future work.
\acksection
This work is partially supported by
funds of the Federal Ministry of Food and Agriculture (BMEL) based on a decision of the Parliament of the Federal Republic of Germany. The Federal Office for Agriculture and Food (BLE) provides coordinating support for artificial intelligence (AI) in agriculture as funding organisation, grant number \textbf{2820KI012}.

\bibliographystyle{apalike}
\bibliography{manual_refs2}

\newpage
\appendix
\section{Representational Similarity}
\label{apx:repsim_measures}
Here, we give an overview of the similarity measures we use and their invariances as well as preprocessing that is applied to the representations in some cases.

\subsection{Invariances}
The following three invariances are relevant for our work:
\begin{itemize}
    \item Invariance to orthogonal transformation (OT) of a representational similarity measure means that $m(\R, \Rp) = m(\R\bm{Q}, \Rp\bm{Q'})$ for any $\bm{Q}, \bm{Q'}$ from the orthogonal group $\mathcal{O}(D)$.
    \item Invariance to isotropic scaling (IS) means $m(\R, \Rp) = m(a\R, b\Rp)$ for any $a,b\in \Real^{+}$.
    \item Invariance to translation (TR) means $m(\R, \Rp) = m(\R + \bm{1}_N\bm{c}^{\transp}, \Rp+ \bm{1}_N\bm{d}^{\transp})$ for any $\bm{c}, \bm{d} \in \Real^D$ with $\bm{1}_N$ being a vector of $N$ ones.
\end{itemize}

\subsection{Preprocessing}
Preprocessing of representations allows us to augment the natural invariances of a similarity measures by eliminating differences that the measure is not invariant to.
Centering columns, i.e., setting the mean of neurons to zero, adds invariance to translation as it removes the same for all representations.
The centered representation is defined as $\widetilde{\R} = \bm{H}\R$ with $\bm{H}=\bm{I}_N - \frac{1}{N} \bm{1}\bm{1}^{\transp}$.
To eliminate scale differences of representations, we can normalize them by their Frobenius norm: $\widetilde{\R} = \R / \|\R\|_F$.
This adds IS invariance.

Finally, some similarity measures require that representations have equal dimensionality.
In those cases, we zero-pad the representation with lower dimension.

\subsection{Representational Similarity Measures}
\paragraph{Orthogonal Procrustes.}
This measure aims to find an orthogonal transformation such that representations are optimally aligned in terms of minimizing the norm of their difference.
Formally:
\begin{equation}
    m_{\mathrm{OP}}(\R,\Rp) = \min_{\bm{Q} \in \mathcal{O}(D)} \|\R \bm{Q} - \Rp \|_F,
\end{equation}
where $\mathcal{O}(D)$ is the orthogonal group of $D$-dimensional matrices and $\|\cdot\|_F$ is the Frobenius norm.
Orthogonal Procrustes is invariant to orthogonal transformations only, but, for our experiments, we add IS invariance by normalizing the representations.
For the experiment with OT, IS, and TR invariance, we first center the representations and then normalize them.

\paragraph{Aligned Cosine Similarity.}
Aligned Cosine Similarity \citep{hamilton_diachronic_2018} is similar to Orthogonal Procrustes in that first an optimal orthogonal alignment of the representation is computed.
Then, the cosine similarity $\operatorname{cossim}(\bm{x}, \bm{y}) = \frac{\bm{x}^\transp\bm{y}}{\|\bm{x}\|_2 \|\bm{y}\|_2}$ is computed between corresponding rows of the representations and aggregated:
\begin{equation}
    m_{\mathrm{ACS}}(\R,\Rp) = \frac{1}{N}\sum_{i=1}^N\operatorname{cossim}(\left(\R\bm{Q}^*\right)_i, \Rp_i)
\end{equation}
with $\bm{Q}^*=\arg\min_{\bm{Q} \in \mathcal{O}(D)} \|\R \bm{Q} - \Rp \|_F$ being the optimal alignment.
Aligned Cosine Similarity is invariant to orthogonal transformations and isotropic scaling.
For the experiment with OT, IS, and TR invariance, we first center the representations.

\paragraph{Norm RSM-Difference.}
This measure uses so-called \emph{representational similarity matrices} (RSMs), that capture the pairwise similarities within one representation matrix---different ways to compute those similarities are possible.
The RSMs of both representations are then compared by taking the norm of their difference.
Formally, let $s: \Real^D \times \Real^D \mapsto \Real$ be a similarity (distance) function between vectors such as cosine similarity.
The choice of $s$ determines the invariances of this measure: cosine similarity is invariant to isotropic scaling and orthogonal transformation, whereas Euclidean distance is invariant to orthogonal transformations and translations.
The matrix of pairwise similarities (the RSM) $\Sm \in \Real^{N \times N}$ between the $N$ representations of $\R$ is then defined elementwise as $\Sm_{i,j} = s(\R_i, \R_j)$.
The similarity between two representations $\R, \Rp$ is computed as \citep{shahbazi_using_2021,yin_dimensionality_2018}:
\begin{equation}
    m_{\mathrm{Norm}}(\R,\Rp) = \|\Sm - \Smp\|_F.
\end{equation}
For the experiment with OT and IS invariance, we use cosine similarity as the similarity function.
For the other experiment, we normalize the representations and use Euclidean distance as the similarity function.

\paragraph{Jaccard Similarity.}
Jaccard Similarity measures the similarity of the nearest neighbors in the representation space \citep{schumacher_effects_2021,wang_towards_2020}.
Let $\mathcal{N}_{\R}^k(i)$ be the set of $k$ nearest neighbors of representation $\R_i$ with respect to a similarity function $s$.
Then, representational similarity is computed as follows:
\begin{equation}
    m_{\mathrm{Jac}}(\R,\Rp) = \frac{1}{N}\sum_{i=1}^N \frac{\left|\mathcal{N}_{\R}^k(i)\cap \mathcal{N}_{\Rp}^k(i)\right|}{\left|\mathcal{N}_{\R}^k(i)\cup \mathcal{N}_{\Rp}^k(i)\right|}.
\end{equation}
Again, invariances depend on the choice of the similarity function $s$. 
We always use cosine similarity.
To add TR invariance, we center the representations first.

\paragraph{Representational Similarity Analysis.}
Representational Similarity Analysis (RSA) \citep{kriegeskorte_representational_2008} uses RSMs similar to Norm RSM-Difference.
Given two similarity functions, $s_{\mathrm{inner}}$ and $s_{\mathrm{outer}}$, first RSMs are computed with $s_{\mathrm{inner}}$. 
Then the RSMs are compared with $s_{\mathrm{outer}}$ by first vectorizing the lower triangle of the RSMs (because they are symmetric) and then applying the similarity function:
\begin{equation}
    m_{\mathrm{RSA}}(\R,\Rp) = s_{\mathrm{outer}}(\operatorname{vec}(\Sm), \operatorname{vec}(\Smp)).
\end{equation}
Again, the invariances are decided by the similarity functions.
For the experiment with OT and IS invariance, we use Pearson correlation as the inner similarity function and Spearman correlation as the outer one.
For the other experiment, we normalize the representations, then use Euclidean distance as the inner similarity function, and finally use Spearman correlation as the outer similarity function.

\paragraph{Centered Kernel Alignment.}
Centered Kernel Alignment (CKA) \citep{cortes_algorithms_2012,cristianini_kernel-target_2001,kornblith_similarity_2019} is a measure that uses the Hilbert-Schmidt Independence Criterion (HSIC) \citep{gretton_measuring_2005}, which can be used to test for statistical independence of two sets of random variables.
HSIC operates on RSMs $\Sm, \Smp$ and is defined as $\operatorname{HSIC}(\Sm, \Smp)=\frac{1}{(N-1)^2} \tr(\Sm \bm{H} \Smp \bm{H})$ with $\bm{H}=\bm{I}_N - \frac{1}{N} \bm{1}\bm{1}^{\transp}$ a centering matrix.
HSIC is divided by a normalization term to achieve IS invariance in addition to OT and TR invariance:
\begin{equation}
    m_{\mathrm{CKA}}(\R,\Rp) = \frac{\operatorname{HSIC}(\Sm, \Smp)}{\sqrt{\operatorname{HSIC}(\Sm, \Sm)\operatorname{HSIC}(\Smp, \Smp)}}.
\end{equation}
CKA requires that representations are preprocessed to have mean-centered columns.
The RSMs are typically computed with the linear kernel, i.e., $\Sm=\R\R^{\transp}$, although other options are possible.
We only use this linear version of CKA.

\section{Prompt Style}\label{apx:prompts}
For Winogrande, we prompt the model in the following style:
\begin{Verbatim}[fontsize=\small]
    Fill in the _ in the below sentence:
    Sentence: Sarah was a much better surgeon than Maria so _ always got the easier cases.
    Option 1: Sarah
    Option 2: Maria
    Does _ in the sentence above refer to Option 1 or 2?
    Answer: Option
\end{Verbatim}
The sentence and the options differ from input to input.
For HumanEval, we follow the prompting style used in the original paper for code completion:
\begin{Verbatim}[fontsize=\small]
    def max_element(l: list):
        """Return maximum element in the list.
        >>> max_element([1, 2, 3])
        3
        >>> max_element([5, 3, -5, 2, -3, 3, 9, 0, 123, 1, -10])
        123
        """
\end{Verbatim}

\section{Additional Experiment Results} \label{apx:additional_results}
\begin{figure}
    \centering
    \begin{subfigure}{\textwidth}
        \centering
        \includegraphics[width=\textwidth]{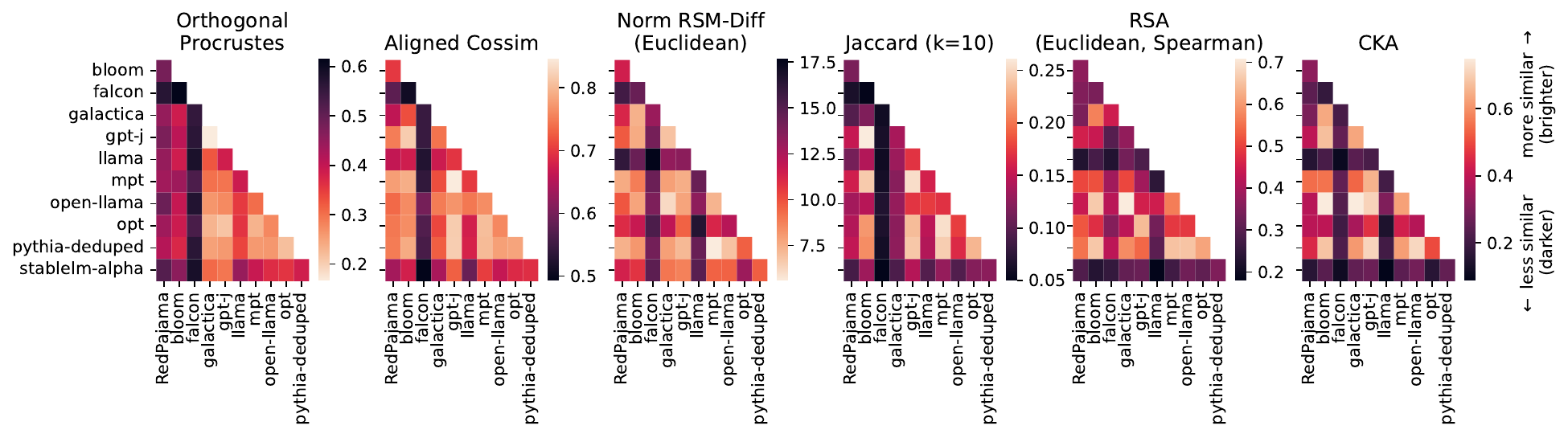}
        \caption{Representational similarity on Winogrande.
        Similar to the other experiment, models show discrepancies and the patterns are different between different similarity measures.
        }
    \label{fig:repsim_otistr}
    \end{subfigure}
    \begin{subfigure}{\textwidth}
        \centering
        \includegraphics[width=\textwidth]{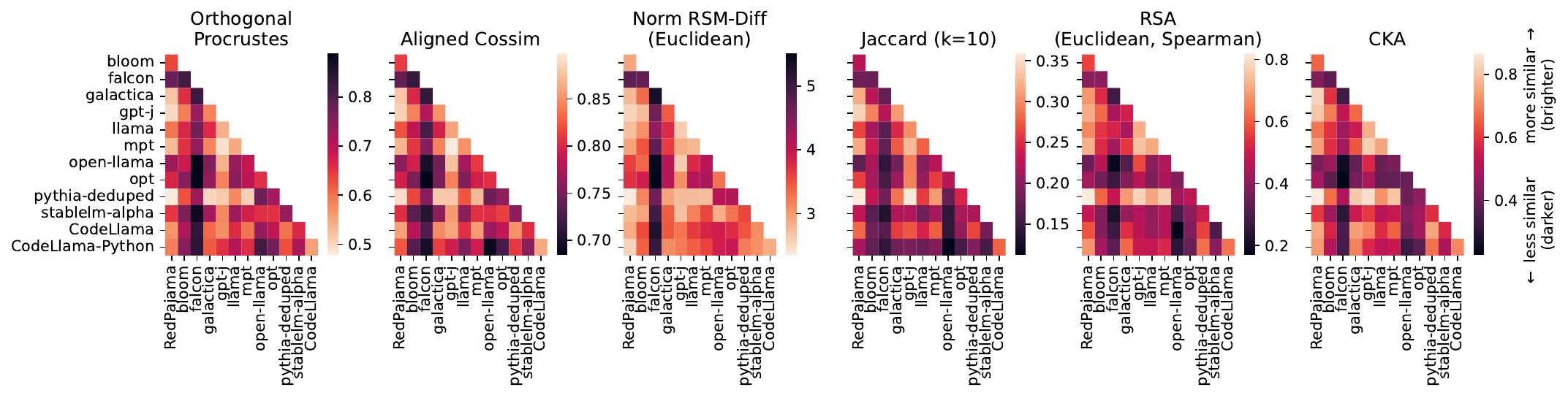}
        \caption{Representational similarity on HumanEval.
        }
        \label{fig:repsim_humaneval_otistr}
    \end{subfigure}
    \caption{Representational similarity on Winogrande (top) and HumanEval (bottom) with OT, IS, and TR invariance.
        Bright colors show the most similar models, dark colors the most dissimilar ones.
    }
    \label{fig:additional_results}
\end{figure}

Additional results for OT, IS, and TR invariance are shown in \Cref{fig:additional_results}.

\section{Code and Data Availability}\label{apx:code}
Code is available at \href{https://github.com/mklabunde/llm_repsim}{https://github.com/mklabunde/llm\_repsim}.
Data is available at \href{https://doi.org/10.5281/zenodo.8411089}{https://doi.org/10.5281/zenodo.8411089}.

\section{Limitations}
Our experiments have several limitations.

First, the datasets we use are limited in size.
The number of samples is lower than the number of dimensions, which may allow overfitting of alignment-based measures like Orthogonal Procrustes and Aligned Cosine Similarity.
Thus, their scores may be overestimated compared to when applied on larger sets of representations.
In future work, additional data---not necessarily from benchmarks---can be used to provide additional evidence.

Second, we only studied models of a specific size.
It is unclear, to what extent these patterns generalize to other models.

Further, we use a specific prompting format, which may influence model similarity.
Importance of prompts for model outputs was demonstrated in prior work \citep{kojima_large_2023,wei_chain--thought_2023}, but has not been studied in detail for representations.

\end{document}